\documentclass[11pt, a4paper, logo, twocolumn, copyright]{googledeepmind}

\usepackage[authoryear, sort&compress, round]{natbib}
\providecommand{\wt}{1h-walk VQA}

\title{Perception Test 2025: Challenge Summary and a Unified VQA Extension}

\correspondingauthor{viorica@google.com}


\author[1]{Joseph Heyward}
\author[1]{Nikhil Parthasarathy}
\author[4]{Tyler Zhu}
\author[1]{Aravindh Mahendran}
\author[1]{Jo\~ao Carreira}
\author[1,2]{Dima Damen}
\author[1,3]{Andrew Zisserman}
\author[1]{Viorica P\u atr\u aucean}

\affil[1]{Google DeepMind}
\affil[2]{University of Bristol}
\affil[3]{University of Oxford}
\affil[4]{Princeton University}

\begin{abstract}
The Third Perception Test challenge was organised as a full-day workshop alongside the IEEE/CVF International Conference on Computer Vision (ICCV) 2025. Its primary goal is to benchmark state-of-the-art video models and measure the progress in multimodal perception. This year, the workshop featured 2 guest tracks as well: KiVA (an image understanding challenge) and Physic-IQ (a video generation challenge).
In this report, we summarise results from the main Perception Test challenge, detailing both the existing tasks as well as novel additions to the benchmark. In this iteration, we placed an emphasis on \textit{task unification}, as this poses a more challenging test for current SOTA multimodal models. The challenge included five consolidated tracks: unified video QA, unified object and point tracking, unified action and sound Localisation, grounded video QA, and hour-long video QA, alongside an analysis and interpretability track that is still open for submissions. Notably, the unified video QA track introduced a novel subset that reformulates traditional perception tasks (such as point tracking and temporal action localisation) as multiple-choice video QA questions that video-language models can natively tackle. The unified object and point tracking merged the original object tracking and point tracking tasks, whereas the unified action and sound localisation merged the original temporal action localisation and temporal sound localisation tracks. Accordingly, we required competitors to use unified approaches rather than engineered pipelines with task-specific models. By proposing such a unified challenge, Perception Test 2025 highlights the significant difficulties existing models face when tackling diverse perception tasks through unified interfaces.     
\end{abstract}

\begin{document}

\maketitle

\section{Introduction}
The quest for a general perception model, a research target that seemed completely out of reach just 3 years ago, seems to be on the horizon due to the impressive performance achieved by large multimodal models (LMMs) like Gemini, GPT, Qwen, and PLM. These models have continued to improve on numerous benchmarks becoming increasingly competitive with human baselines. Our team introduced the Perception Test benchmark~\citep{patraucean2023perception} to comprehensively measure the performance of multimodal video models on a variety of perception tasks and across modalities. Since its release, we have organised public challenges in conjunction with top computer vision venues to document the progress towards general perception models. 

\noindent Given the rapid progress of LMMs, the 2025 edition (organised as an ICCV 2025 workshop) placed an emphasis on \textit{task unification}, introducing several multi-task challenges,  each requiring a corresponding single or "unified" model as a solution.

\begin{figure}
    \centering
    \includegraphics[width=\linewidth]{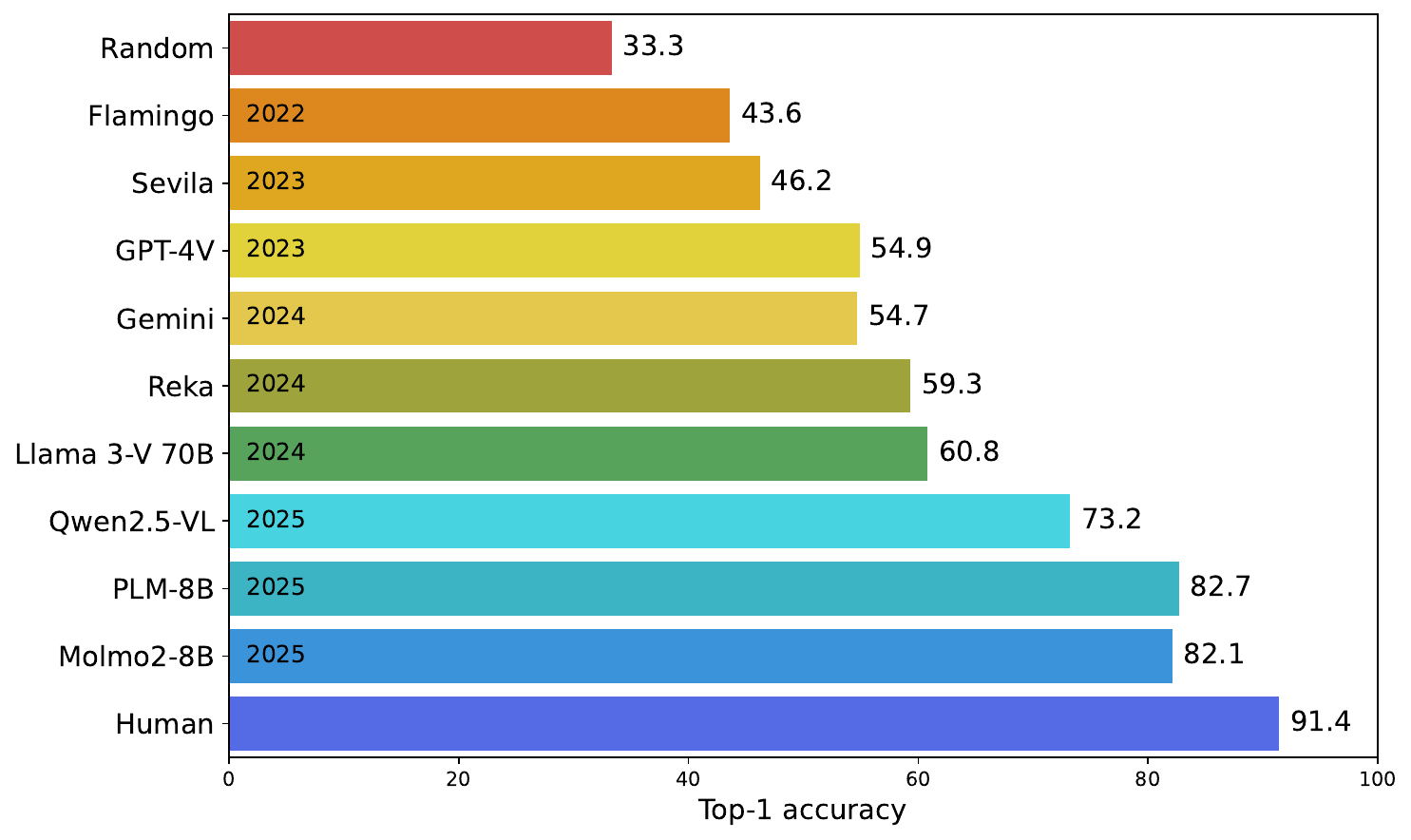}
    \caption{Top-1 accuracy of various VLMs at the moment of their first release compared to human baseline on the Perception Test multiple-choice video QA task. We include the results published by models' authors where available, otherwise we ran the models independently at the moment of their first release (GPT-4V, SeViLA, Flamingo).}
    \label{fig:sotahuman}
\end{figure}

\noindent \textbf{Challenge tracks:} The 2025 challenge\footnote{\url{https://perception-test-challenge.github.io/}} had five tracks: joint object and point tracking, joint action and sound temporal localisation, grounded videoQA, unified multiple-choice videoQA, and hour-long videoQA. The first three tracks relied on videos and annotations available in the original Perception Test benchmark, with the first two tracks providing a unifying interface for previously separated tasks, e.g. the point tracking and object tracking were unified under the same tracking interface. The unified videoQA track used videos from the Perception Test benchmark and added a language interface for different types of annotations (point tracks, action segments). The hour-long track relied on the Walking Tours dataset introduced in the 2024 edition~\citep{heyward2024perceptiontest2024challenge}. We describe in the next sections the setup, metrics, and results in each track.

\noindent \textbf{Challenge setup:} We relied on the open-source eval.ai platform to set up the different challenge tracks. Each track had 2 phases (validation and test). For each submission, the participants had to indicate the evaluation mode (fine-tuning, few-shot, or zero-shot evaluation). For test submissions, the participants were required to also upload a short report describing their method (architecture, pre-training datasets and tasks, etc.).
The validation phase served as a sanity check for participants' submission pipelines. The number of submissions for the validation phase was not limited. The test set was made available 1 month before the submission deadline. 
For the test phase, the limit was set to 2 submissions per day, 30 submissions in total. 
Only the results made public on the test leaderboard were considered for the competition. 

For the joint tracks, the requirement was for competing models to be able to solve both types of queries simultaneously. E.g. a solution where a given pre-trained model is fine-tuned in sequence on object tracking and separately on point tracking, would not be eligible.

\section{Unified VQA: A novel addition to the Perception Test}
\label{sec:uvqa}

\begin{figure*}
    \centering
    \includegraphics[width=0.95\textwidth]{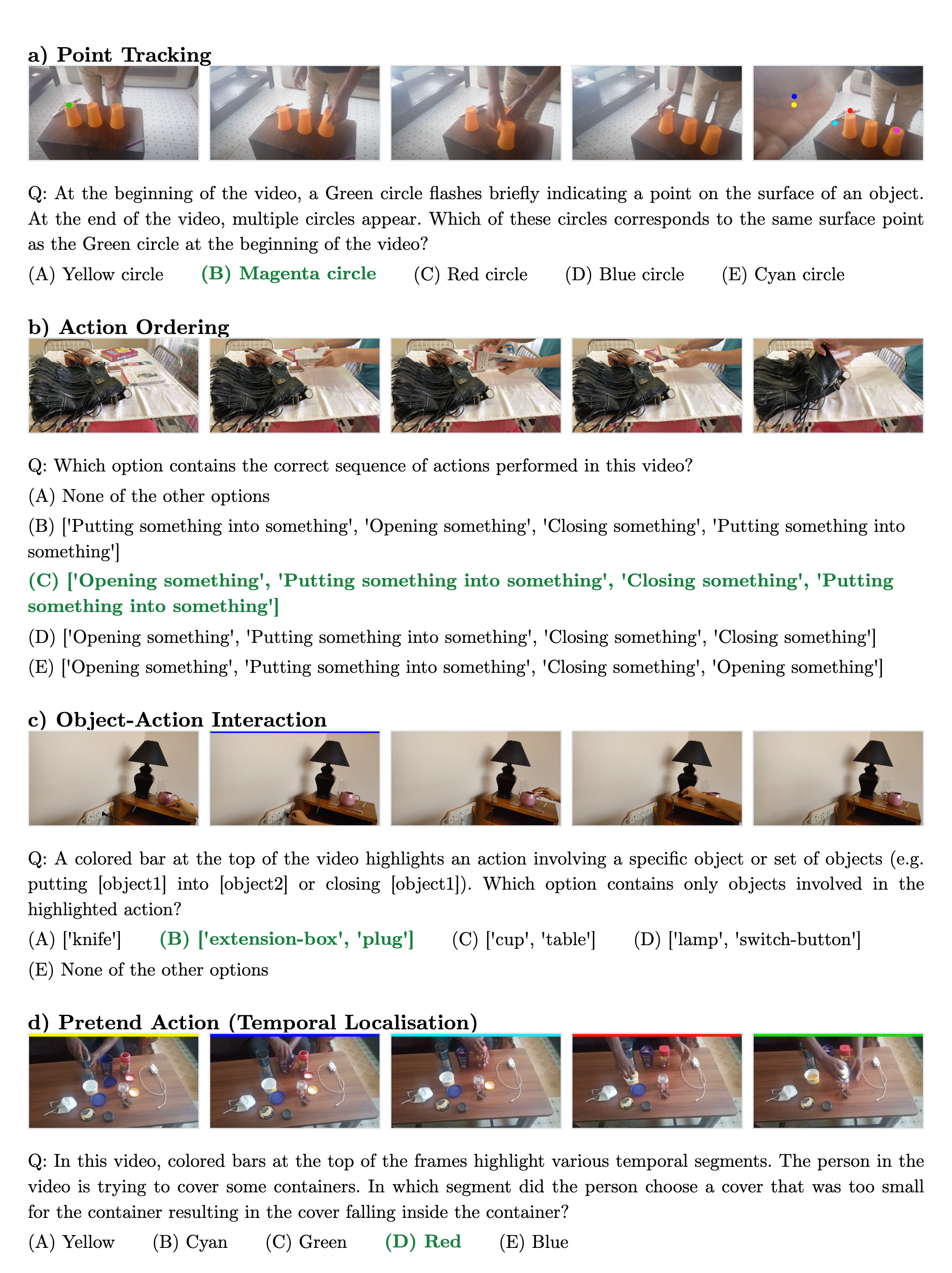}
    \caption{An example of each of the 4 new question types in the unified mc-VQA set (labeled a-d). We show sampled frames from the videos (some with visually in-painted annotations) that help define the task. The model must then choose the correct answer (highlighted in green) from the 5 options.}
    \label{fig:unified-example}
\end{figure*}

In the age of Vision-Language Models (VLMs), language has become a universal interface for probing the visual understanding of state-of-the-art multimodal models. While the original Perception Test benchmark \citep{patraucean2023perception} contained a large set of mc-VQA questions that has become a standard for VLM evaluations, several core capabilities—notably point tracking and temporal action localization—required separate benchmarks, each with their own disjoint evaluation procedures. To address this, we introduce a \textbf{Unified mc-VQA} set. This framework leverages visual in-painting and temporal markers (e.g., colored bars and flashing indicators) to re-cast traditionally complex localization and tracking tasks as robust VQA probes. It is worth noting that while our efforts focus on benchmarking spatiotemporal understanding in video, there have been similar image understanding benchmarking efforts \citep{fu2024blink, chen2024spatialvlm} that cast traditional spatial understanding tasks (e.g. visual correspondence and depth estimation) as VQA.

The unified set augments the original 11,528 samples (3-choice) with \textbf{1,842 new questions} utilizing a 5-choice format to mitigate shortcut solutions and increase the difficulty of the reasoning task. These new questions are distributed to test specific facets of spatiotemporal reasoning:
\begin{itemize}
    \item \textbf{Point Tracking (652 questions):} Uses a brief ``Green Flash'' to indicate a specific surface point; the model must identify that same point among multiple colored circle options at the end of the video. Critically, due to the frequent presence of identical instances of the same object, accurate point tracking is essential to correctly answer these questions (see Fig~\ref{fig:unified-example}).
    \item \textbf{Action Ordering (676 questions):} Requires models to identify the correct linear sequence of performed actions, testing long-term temporal modeling.
    \item \textbf{Object-Action Interaction (245 questions):} Targets the ``binding problem'' by requiring models to identify the specific set of objects involved in a highlighted temporal segment from a longer video that showcases a number of unique object interactions.
    \item \textbf{Pretend Action (269 questions):} Probes complex intent understanding as well as temporal action localisation by asking the model to identify segments where an action is ``fake'' or intentionally incorrect (indicated by highlighted temporal segment markers).
\end{itemize}

By grounding spatial and temporal queries directly in the visual stream via in-painting, we bypass the need for complex task decoders and focus instead on the model's ability to integrate visual prompts with high-level reasoning. This unified structure ensures that Perception Test remains a challenging probe for SoTA VLMs. We present an example of each question type from the unified mc-VQA point tracking subset in Figure~\ref{fig:unified-example}. 

\section{Overall challenge summary}  

\begin{figure}
    \centering
    \includegraphics[width=\linewidth]{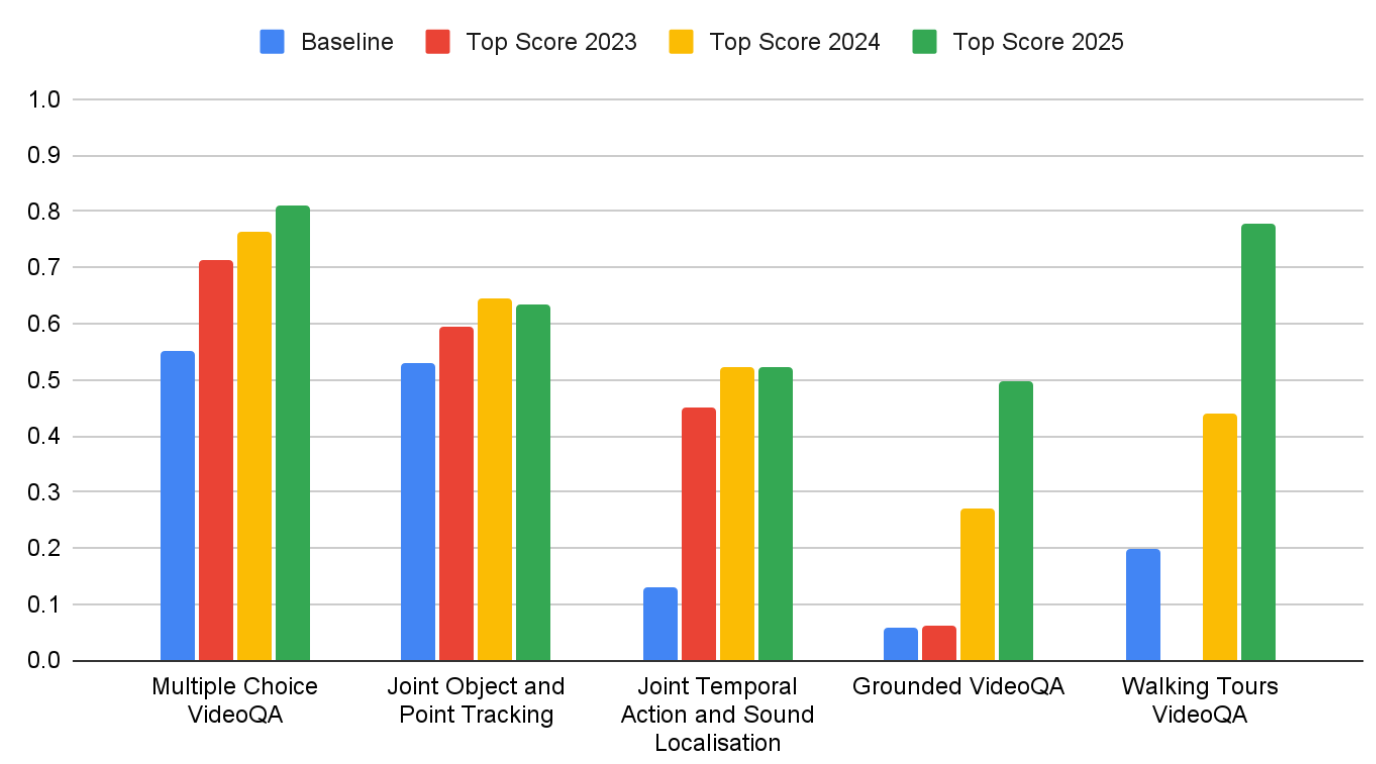}
    \caption{Per-track performance improvement compared to baselines and compared to best models from previous years.}
    \label{fig:summary2025}
\end{figure}

We received 450 submissions from over 100 different teams across all five tracks. We awarded 2 prizes per track (best and runner-up) to submissions that obtained the best (and second best) results in the test leaderboard, with prizes totalling 47k USD (up from 20k EUR in 2024). As it can be observed in Figure~\ref{fig:summary2025}, for the tracks with language interface (unified videoQA, hour-long video QA, grounded videoQA), the top performing models improved compared to the winning models from last year, with large jumps in grounded videoQA and hour-long videoQA (performance almost doubled compared to 2024). This shows the impressive progress that video-language models have made this last year in terms of being able to process longer videos and localise and track objects. 

For the merged tracks without language interface (object \& point tracking, action \& sound localisation), the top performance this year didn't exceed the average of the top winners in the corresponding tracks from last year, e.g. in 2024, the top performing model in point tracking obtained 0.47 AJ and in object tracking 0.81 IoU. The top unified model this year obtained 0.62. While promising as these multi-task models are more capable, our benchmark shows that there is still good room for improving their performance. 

\begin{figure}
    \centering
    \includegraphics[width=\linewidth]{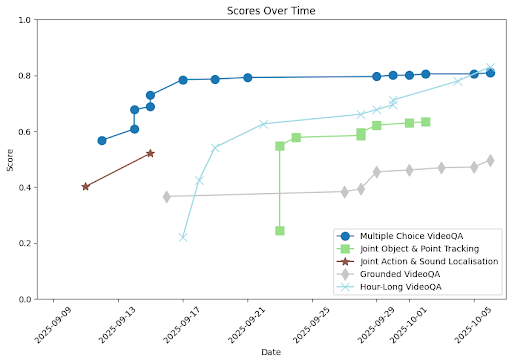}
    \caption{Per-task performance improvement of top models during the 2025 test submission phase.}
    \label{fig:pertask}
\end{figure}

Overall, the improvement from last year reflects the current focus of the community, i.e. improving the performance of video language models. Figure~\ref{fig:pertask} shows the evolution of the top-performing models during the test submission phase of this year's edition for each track. The reports of the winning submissions are available on the workshop website.

\section{Challenge Tracks, Results, Awards}
In the following we describe each track and the performance achieved in the challenge. For the technical report per team, including winners' affiliations and names, please refer to the workshop website: \url{https://perception-test-challenge.github.io/}.

\subsection{Unified multiple-choice video QA}

\noindent \textbf{Task description:} In the unified multiple-choice video question-answering (mc-vQA) task, the model receives, in parallel with the video, a question and three or five possible answers, out of which only one is correct, and the model has to pick one answer. The questions cover four skill areas (Memory, Abstraction, Physics, Semantics) and require different types of reasoning (Descriptive, Explanatory, Predictive, Counterfactual), across video, audio, and text modalities. The questions are also tagged with skills in each area such as: event recall (Memory), object counting (Abstraction), collision (Physics), action recognition (Semantics) and more. 

\noindent \textbf{Dataset:}
We use the set of videos and questions from the original Perception Test multiple-choice video QA task together with the extra set added this year, described in section~\ref{sec:uvqa}. Each video in the dataset has a number of multiple-choice video QA tasks associated, each question having 3 options in the original set and 5 options in the new set; only one option is correct in both sets. Table~\ref{tab:mcqa} provides details about the number of videos and number of questions in this track.

\begin{table}[h]
    \centering
    \small{
    \begin{tabular}{lll}
        \toprule
        Split & Num videos  & Num questions\\ 
        \midrule
        Train & $2184 + 0$ & $7392 + 0$ \\
        \addlinespace
        Validation & $5900 + 189$ & $19140 + 454$ \\
        \addlinespace
        Test & $3525 + 1842$ & $11528 + 1842$ \\
        \bottomrule
    \end{tabular}}
    \caption{Number of videos and questions in the original plus the extra set added this year (2025). For the extra set, we only provided samples in the validation and test splits, to enforce zero-shot evaluation.}
    \label{tab:mcqa}
\end{table}

\noindent \textbf{Metric:} The evaluation metric for this challenge is top-1 accuracy. It is calculated as the percentage of questions where the model's predicted option id (1 out of 3 in the original set and 1 out of 5 in the newly added set) matches the ground truth option id.

\noindent \textbf{Baselines:} We provide baseline results for this task using a random baseline that scores 0.31\% (weighted average between 0.33\% the expected score for a random baseline in the original set, and 0.20\% the random baseline in the new set with 5 options per question), and a frequency-based baseline, which uses at test time the most frequent answer from the train set for each question.

\begin{figure}[t]
    \centering
    \includegraphics[width=.95\linewidth]{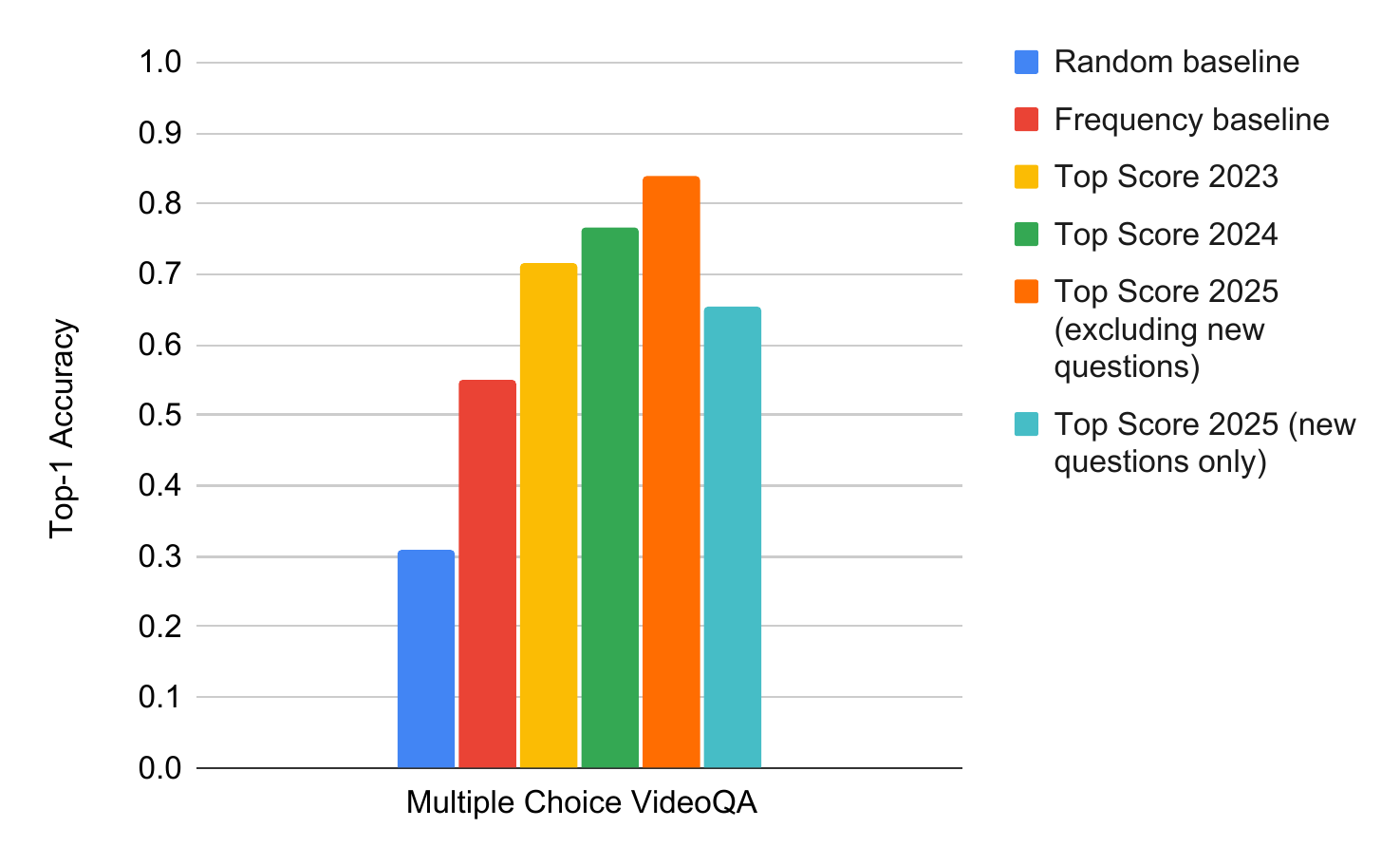}
    \caption{Evolution over time of the best performing models competing in our challenges for the multiple-choice videoQA task compared to random and frequency-based baselines.}
    \label{fig:mcqa}
\end{figure}

\noindent \textbf{Results:}

\begin{table}[h]
    \centering
    \small{
    \begin{tabular}{llc}
    \toprule
       \textbf{Rank} & \textbf{Team name} & \textbf{top-1} \\
       \midrule
       Baseline & Random & 0.314 \\
       Baseline  & Frequency (all-shot) & 0.552 \\
       Runner-up & PCIE\_VQA (fine-tuned) & 0.806 \\
       Best & njust\_kmg\_2 (fine-tuned) & 0.810 \\
       Best & Wind\_Rain\_Tower (zero-shot) & 0.812\\
         \bottomrule
    \end{tabular}}
    \caption{Multiple-choice video QA results.}
    \label{tab:mcqaresults}
\end{table}

Table~\ref{tab:mcqaresults} shows the performance of the top competing models compared to our dummy baselines. In this track, we awarded two best-model awards, as we had 2 submissions with very close scores and different approaches (one fune-tuned, one zero-shot). The zero-shot top model used a simple dual-model framework based on Seed 1.6 Vision~\citep{guo2025seed15vltechnicalreport} and Qwen-VL-Max models~\citep{Qwen2VL}: if the first model failed to generate an answer, the second model would be called. The fine-tuned top model used an ensemble approach formed of finetuned Qwen-2.5VL and APIs of GLM-4.5V and Seed-1.6-Vision. Please check the workshop website for more details on the method included in the submission report.

Figure~\ref{fig:mcqa} shows the performance of the zero-shot best 2025 submission compared to the top results from the previous years. It can be observed that the model made a significant improvement compared to previous years on the original test set (83.7\% accuracy) and its performance on the novel set is also quite strong (65.4\%), indicating strong generalisation capabilities. 

\begin{figure*}[h]
    \centering
    \includegraphics[width=\linewidth]{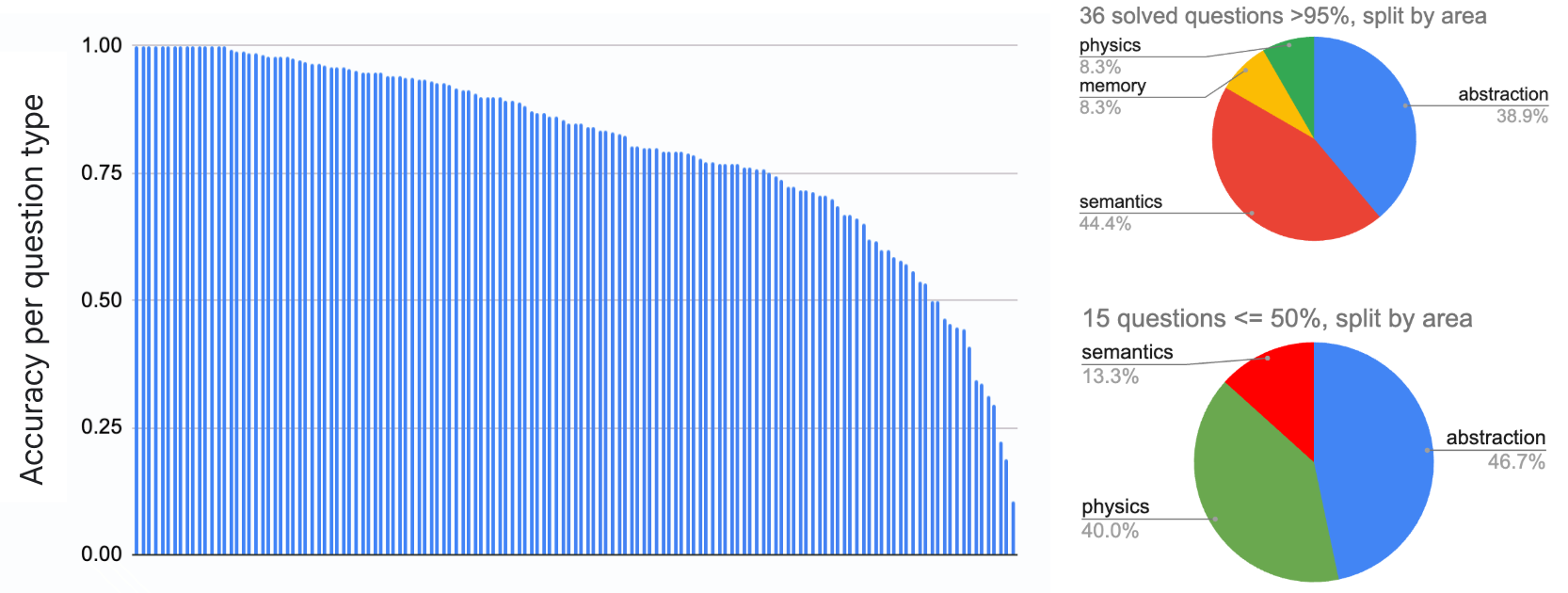}
    \caption{Left: Performance of the top zero-shot model across question types. Right: Distribution across areas of the questions that are quasi-solved (top) and questions that are still very challenging (bottom).}
    \label{fig:statsvqa}
\end{figure*}

In Figure~\ref{fig:statsvqa}, we show the performance of the top zero-shot model across individual questions, and the distribution across areas of the  questions that are quasi-solved (accuracy $>95\%$) and questions that are still very challenging (accuracy $<50\%$). Out of the 135 total unique questions, 27\% of the questions can be considered solved. The most challenging questions are in the Abstraction and Physics areas. The questions in the Memory area do not seem to pose significant challenges to SOTA models, as these models keep all the video frames in context, hence memory tasks become attention over time tasks, which Transformer models can perform very well. 

\subsection{Object \& point tracking}

\noindent \textbf{Task description:} The joint object \& point tracking task has two types of queries:
(1) object query: the model receives a video and a bounding box representing an object, and it is required to track the object throughout the video; (2) point query: the model receives a video and the 2D coordinates of a point, and it is required to track the point throughout the video, also accounting for occlusions.

\noindent \textbf{Dataset:} For this joint task, we combined the datasets used in the separate tracks in the previous years; see details in Table~\ref{tab:all_tracks}.

\begin{table}[h]
\centering
\small{
\begin{tabular}{llrr}
\toprule
\textbf{Query} & \textbf{Split} & \textbf{Num videos} & \textbf{Num tracks} \\ \midrule
Object              & Train          & 2184                & 35373               \\
Object              & Val     & 1000                & 16501               \\
Object              & Test           & 1000                & 16339               \\ \midrule
Point               & Train          & 28                  & 1758                \\
Point               & Val     & 73                  & 4362                \\
Point               & Test           & 44                  & 2527                \\ \bottomrule
\end{tabular}}
\caption{Dataset used for the joint object \& point tracking challenge.}
\label{tab:all_tracks}
\end{table}

\noindent \textbf{Metric:} The evaluation metric for this joint challenge is average Jaccard. This is calculated as the average between the point average Jaccard proposed in TAP-Vid~\citep{doersch2022tapvid} and the object average Jaccard (standard IoU).

\noindent \textbf{Baseline:} We provide baseline results for this task using a dummy static baseline, which always assumes that the point or object is static and visible in all frames. This baseline obtains 0.53 AJ.

\noindent \textbf{Results:}
Table~\ref{tab:point_tracking} shows the results of the top competing models compared to our static dummy baseline. In this track we awarded a runner-up prize and two best model prizes, and the two top submissions obtained close scores using very different approaches. NJUST\_KMG team relied on a pipeline approach combining SAM2~\citep{sam2} and AllTracker~\citep{harley2025alltracker}, whereas Team 30947 used a unified model, based on TAG~\citep{harley2024tag}. Please check the workshop website for more details on the method included in the submission report.

Figure~\ref{fig:summary2025} compares the top-performing model this year against the top submissions from previous years, averaged across the previously separate object and point tracking challenges. It can be observed that the top model this year did not succeed to outperform the best specialised models submitted last year.

\begin{table}[]
    \centering
    \small{
    \begin{tabular}{l|l|c}
    \toprule
       \textbf{Rank} & \textbf{Team name} & \textbf{Avg Jaccard} \\
       \midrule
       Baseline & Dummy static & 0.53 \\
    Runner-up & LC & 0.59 \\
       Best & 30947 & 0.60 \\
       Best & NJUST\_KMG & 0.62 \\
         \bottomrule
    \end{tabular}}
    \caption{Results in the joint object and point tracking challenge.}
    \label{tab:point_tracking}
\end{table}

\subsection{Action \& sound temporal localisation}

\noindent \textbf{Task description:} In the joint action and sound temporal localisation task, the model receives a video and is required to localise and classify the actions and sounds occurring in the video according to predefined sets of classes for actions and sounds. A single model must perform both action and sound localisation tasks.

\noindent \textbf{Dataset:} We use the videos from the Perception Test benchmark and we combine the annotations for action and sound segments, see Table~\ref{tab:tasl} for details. To facilitate experimentation, we also provide features for the video / audio modalities that participants could optionally use for their submissions: video features extracted using TSP~\citep{alwassel2021tsp} and audio features extracted using MMV~\citep{alayrac2020self}. 
\begin{table}[h]
    \centering
    \small{
    \begin{tabular}{llll}
        \toprule
        Split & Num videos & Num actions & Num sounds \\ 
        \midrule
        Train      & 2009            &  13097 & 13289             \\  
        Val & 5359          & 35440 & 35625           \\ 
        Test       & 3233 & 20741 & 21636 \\
        \bottomrule
    \end{tabular}}
    \caption{Dataset used for the action \& sound temporal localisation challenge.}
    \label{tab:tasl}
\end{table}

\noindent \textbf{Metric:} The evaluation metric for this challenge is mean average precision (mAP). This is calculated as the average precision over different classes and IoU thresholds for actions and sounds separately and then averaged using equal weights. For the IoU thresholds in evaluation we use [0.1 $\rightarrow$ 0.5] with 0.1 increments, similar to~\citep{Damen2021TheED}.

\noindent \textbf{Baseline:} The baseline for this task is ActionFormer~\citep{zhang2022actionformer} that we fine-tuned for the set of classes present in our benchmark.

\noindent \textbf{Results:} 
The results of the top competing methods are included in Table~\ref{tab:tal} and are compared against our baseline. The top entry this year was submitted by NJUST--\_KMG Team and uses a multimodal ActionFormer with cross-modal alignment losses. Please check the authors' report on our workshop page for more details. Figure~\ref{fig:summary2025} compares the top-performing model this year against the top submissions from previous years, averaged across the previously separate action and sound temporal localisation challenges.   

\begin{table}[]
    \centering
    \begin{tabular}{l|l|c}
       \textbf{Rank} & \textbf{Team name} & \textbf{mAP} \\
       \hline
       Baseline & ActionFormer & 0.13\\
       Runner-up & SV & 0.50 \\
       Best & NJUST--\_KMG & 0.52\\
         \hline
    \end{tabular}
    \caption{Results in the action \& sound temporal  localisation challenge.}
    \label{tab:tal}
\end{table}

\subsection{Grounded video QA}

\noindent \textbf{Task description:} In the grounded video QA task, the model receives a video and a question/query as input, and it is required to track throughout the video the object(s) that represent the answer to the question.

\noindent \textbf{Dataset:} We use the videos from the Perception Test that have annotations for this task matching the dataset used in previous years; see details in Table~\ref{tab:gvqa}.

\begin{table}[h]
    \centering
    \small{
    \begin{tabular}{lll}
        \toprule
        Split & Num videos & Num questions \\ 
        \midrule
        Train      & 586            &  1859              \\  
        Val & 1545          & 3051            \\ 
        Test       & 932 & 1859 \\
        \bottomrule
    \end{tabular}}
    \caption{Dataset used for the grounded video QA challenge.}
    \label{tab:gvqa}
\end{table}

\noindent \textbf{Metric:} The evaluation metric for this track is HOTA (Higher Order Tracking Accuracy)~\citep{luiten2020IJCV}. It unifies the detection, association, and localization accuracy into a single metric.

\noindent \textbf{Baselines:} We provide a simple baseline that runs MDETR detector~\citep{kamath2021mdetr} on the middle frame of the video using the given question as query, then it keeps the detections static throughout the video.

\noindent \textbf{Results:}
The top-2 results for this track are included in Table~\ref{tab:gqa} compared to our baseline. The top model used a pipeline formed of 3 models: Gemini 2.5Pro for answering the questions, Molmo~\citep{molmo2024} for grounding the answer, and SAM2~\citep{sam2} for tracking the objects referenced in the answer. The runner-up solution managed to solve the task using only 2 models: the Seed 1.6 Vision~\citep{guo2025seed15vltechnicalreport} to answer the question directly as object coordinates, and SAM2~\citep{sam2} for tracking bi-directionally. 

\begin{figure}
    \centering
    \includegraphics[width=\linewidth]{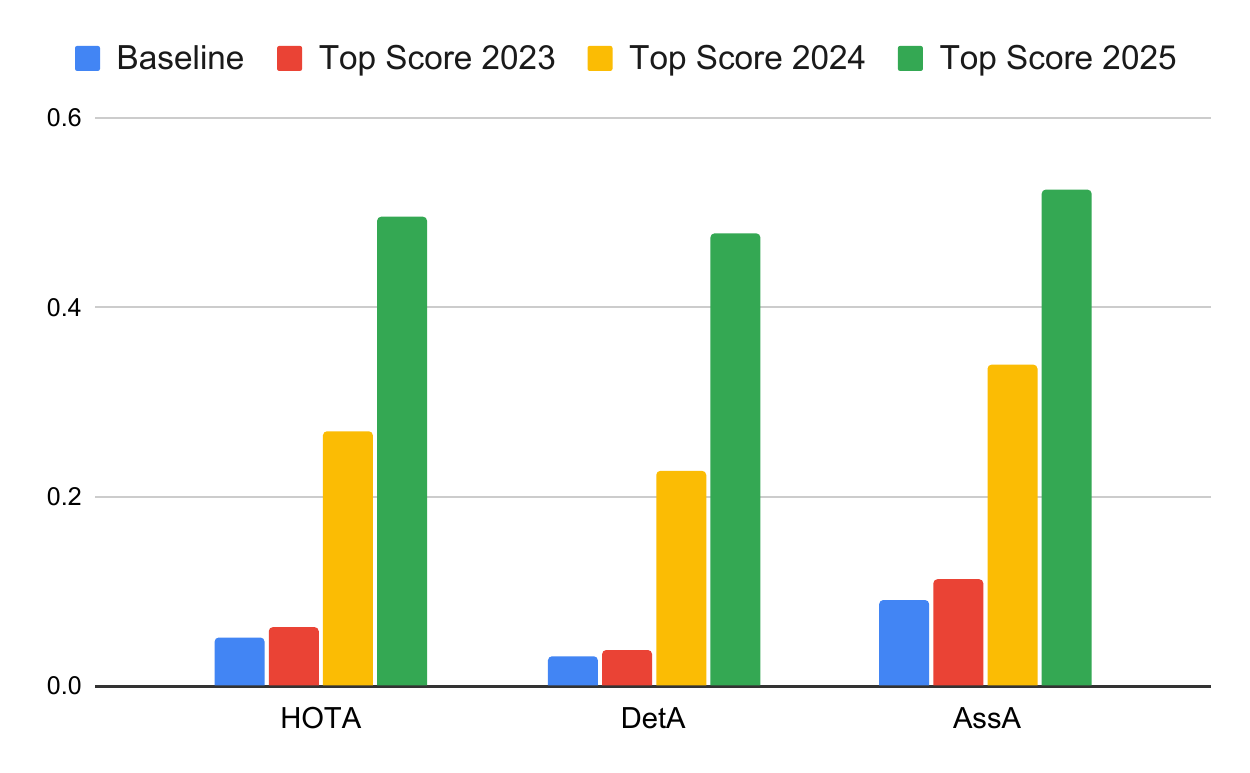}
    \caption{Baseline vs best results over time in terms of overall HOTA, detection, and assignment accuracy for the grounded video QA challenge.}
    \label{fig:hota}
\end{figure}

Figure~\ref{fig:hota} compares the top model this year to the best competitors in previous years, showing an impressive progress, almost doubling the previous performance. These results highlight very clearly the pace of progress in the field: two years ago, the models were struggling to obtain better-than-chance performance, whereas now they provide strong results even for tasks on which they were not trained explicitly.

\begin{table}[h]
    \centering
    \small{
    \begin{tabular}{llc}
    \toprule
       \textbf{Rank} & \textbf{Team name} & \textbf{HOTA} \\
       \midrule
       Baseline & MDETR+static & 0.057 \\
              Runner-up & TutuAI & 0.430 \\
       Best & SGVR@KAIST & 0.499 \\
         \bottomrule
    \end{tabular}}
    \caption{Grounded video question-answering results.}
    \label{tab:gqa}
\end{table}

\subsection{Hour-long video QA}

\noindent \textbf{Task description:} In the hour-long video question-answering task, the model receives a very long video (1h or longer) and questions with five possible answers, out of which only one is correct.

\noindent \textbf{Dataset:} We use the \textit{\wt}\ benchmark introduced in the 2024 edition of the challenge~\citep{heyward2024perceptiontest2024challenge}, which relies on Walking Tours videos to define challenging multi-hop multimodal reasoning tasks in multiple-choice video QA format. This benchmark is small and intended for zero-shot evaluation, no training or fine-tuning data are available; see details in Table~\ref{tab:hvqa}. 
\begin{table}[]
    \centering
    \small{
    \begin{tabular}{l|r|r}
    \toprule
      \textbf{Split} & \textbf{\# videos} & \textbf{\# questions} \\
      \midrule
      Train & -  & - \\
      Validation  & 3 & 11 \\
      Test  & 7 & 59 \\
      \bottomrule 
    \end{tabular}}
    \caption{Splits in \textit{\wt}\ benchmark used for the hour-long video QA task.}
    \label{tab:hvqa}
\end{table}

\noindent \textbf{Metric:} The evaluation metric for this challenge is top-1 accuracy. It is calculated as the percentage of questions where the model's predicted option id (1 out of 5) matches the ground truth option id.

\noindent \textbf{Baselines:} We consider a dummy random baseline for this task, and a zero-shot human baseline, which obtained 99.64\%~\citep{heyward2024perceptiontest2024challenge}.

\noindent \textbf{Results:}
The top-2 results for this track are included in Table~\ref{tab:hvqawin}, compared to the above baselines. Both these submissions relied on Gemini 2.5 Pro to caption short segments of the video, together with chain-of-thought reasoning. together; please check the workshop website for more details. The improvement this year compared to last year is impressive (Figure~\ref{fig:summary2025}), showing that current SOTA large multimodal models have very strong perception capabilities even when dealing with very long videos.

\begin{table}[]
    \centering
    \small{
    \begin{tabular}{l|l|c}
       \textbf{Rank} & \textbf{Team name} & \textbf{HOTA} \\
       \hline
       Baseline & Random & 0.203 \\
       Baseline & Human & \textbf{0.996} \\
              Runner-up & Oair\_Hunter & 0.661 \\
       Best & NJUST—\_KMG (ge) & 0.780 \\
         \hline
    \end{tabular}}
    \caption{Hour-long video question-answering results.}
    \label{tab:hvqawin}
\end{table}

\section{Discussion}

The Third Perception Test challenge attracted a large number of submissions from over hundred teams across all tracks. This edition focused on unifying tasks, to go beyond specialised single-task models (e.g. joint object and point tracking, joint action and sound localisation). In addition, we introduced a novel task set based on videos from the original benchmark that provides a language interface to non-language tasks using inpainting. Overall, we observed a clear improvement in performance for the tracks with language interface, especially in the grounded video QA track and the hour-long video QA track, where the performance almost doubled compared to last year, showing how fast the field is moving, with models getting more and more close to human performance.

\subsection*{Acknowledgements} We would like to thank Google DeepMind for providing the funding for the awards and swag. Special thanks to the Eval AI team for their support while running the challenges.

\bibliography{main}

\end{document}